\pdfoutput=1
\documentclass[11pt]{article}

\usepackage[]{ACL2023}

\usepackage{times}
\usepackage{latexsym}
\usepackage{graphicx}

\usepackage[T1]{fontenc}
\usepackage[utf8]{inputenc}

\usepackage{microtype}

\usepackage{inconsolata}

\usepackage{booktabs}

\usepackage{lipsum}

\newcommand\blfootnote[1]{%
  \begingroup
  \renewcommand\thefootnote{}\footnote{#1}%
  \addtocounter{footnote}{-1}%
  \endgroup
}

\title{Training an NLP Scholar at a Small Liberal Arts College: A Backwards Designed Course Proposal}

\author{Grusha Prasad\textsuperscript{*} \\
  Department of Computer Science \\
  Colgate University \\
  \texttt{gprasad@colgate.edu} \\\And
  Forrest Davis\textsuperscript{*} \\
  Department of Computer Science \\
  Colgate University \\
  \texttt{fdavis@colgate.edu} \\}

\begin{document}
\maketitle

\begin{abstract}

The rapid growth in natural language processing (NLP) over the last couple years
has generated student interest and excitement in learning more about the field. In this paper, we present two types of students that NLP courses might want to train. First, an ``NLP engineer'' who is able to flexibly design, build and apply new technologies in NLP for a wide range of tasks. Second, an ``NLP scholar'' who is able to pose, refine and answer questions in NLP and how it relates to the society, while also learning to effectively communicate these answers to a broader audience. While these two types of skills are not mutually exclusive --- NLP engineers should be able to think critically, and NLP scholars should be able to build systems --- we think that courses can differ in the balance of these skills. As educators at Small Liberal Arts Colleges, the strengths of our students and our institution favors an approach that is better suited to train NLP scholars. In this paper we articulate what kinds of skills an NLP scholar should have, and then adopt a backwards design to propose course components that can aid the acquisition of these skills. 
%In reflecting on the strengths of our students and our institution, a small
% liberal arts college, we designed our course around the type of student we want
% to train -- the NLP Scholar. A student focused on developing the skills to pose,
% refine, and answer questions in NLP. Utilizing backwards design, we developed a
% course that centers the skills we believe are necessary to train an NLP scholar. [THIS IS SUCH TRASH BUT WANTED SOME WORDS HERE]
\end{abstract}

\section{Introduction}
\blfootnote{\textsuperscript{*}Equal contribution.}

Natural language processing (NLP) as an academic discipline has undergone a rapid expansion in the last decade. Moreover, the feverishness around emerging technologies from NLP influences what students want to learn in their CS education. They want courses that train them in the tools and techniques of both NLP and Machine Learning more broadly. As educators, we are faced with an exciting challenge -- how do we effectively train students to engage productively with our field? 

In this paper, we differentiate between two kinds of desired learning outcomes for an NLP course: students should be able to (1) build and use new technologies, and (2) pose, refine and answer questions in NLP. While NLP courses might seek to achieve both of these outcomes, the relative emphasis placed on each of them can differ across courses. Some courses might place a larger emphasis on (1), and present students with many opportunities to program and develop substantial projects, often with a focus on working with state-of-the-art approaches.  In contrast, some courses might place a larger empahsis on (2) and present students with opportunities to engage critically with NLP tools and techniques, often in the service of answering questions and facing challenges that bridge disciplinary boundaries. We will refer to the type of student the former course is designed to train as an \textbf{``NLP engineer''}, and the student for the latter course as an \textbf{``NLP scholar''}.

As educators in a Computer Science Department at Colgate University (a Small Liberal Arts College; SLAC),  where the curriculum structure requires students to take a wider range of non-CS classes with the goal of cultivating critical thinking, we argue that we are better placed to train NLP scholars, than we are to train NLP engineers. In this paper we present a design for an upper level NLP course intended to train NLP scholars. 

Using a backward design, we start by articulating what are the skills that an NLP scholar should have, and propose a set of course principles/tenets that can facilitate the acquisition of these skills. Then, drawing on our previous experience teaching Applied Machine Learning and Natural Langauge Processing at our institution, 
%a Small Liberal Arts College (SLAC), an environment we argue is ideal to train NLP Scholars, 
we propose a course that is structured around a capstone project. Concretely, our paper makes three contributions: 

\begin{enumerate}
    \item We make explicit the link between the kinds of students we want to train, the desired skills we want the students to have, and the course structure and content. 
    \item We propose a course structure designed around a capstone experience which has six different course components and relate each of these components to concrete concepts and skills they are designed to help with. 
    \item We present preliminary materials and evaluation rubrics for several of these components.
\end{enumerate}

%We hope that highlighting some of the decisions we can make when designing our NLP courses and tying them back to concrete learning objectives can facilitate further discussion around the types of pedagogies we can adopt to train both NLP scholars and engineers.

The paper is structured as follows: First, in \S\ref{sec:positionality},  we take critical stock of our students, their experiences, and their relation to computer science, asking
ourselves how can we build on the strengths of our students and our institution. 
%We summarize our positionality in \S\ref{sec:positionality}.
Then, in \S\ref{sec:scholar}, we present skills we think an NLP scholar should have, and the principles we think a course designed to train NLP should adopt. Finally, in \S\ref{sec:course}, we detail our different course components and discuss how they relate to our overall learning objectives.\footnote{A github repository with links to final project templates and our toolkit can be found here: \url{https://github.com/forrestdavis/NLPScholar}.}

\section{Teaching NLP as an upper level course at a Small Liberal Arts College}\label{sec:positionality}
Small Liberal Arts College (SLACs) are undergraduate focused institutions that emphasize, among other things, drawing connections between different fields \cite{king2007liberal}. Therefore, in these institutions there is more emphasis on students taking classes in different disciplines. A consequence of this is that CS departments in liberal arts colleges tend to have fewer required courses and curricula with flatter prerequisite structures \cite{teresco2022cs}. In this section we introduce the CS curriculum structure and class room dynamics at Colgate University, a SLAC in the US that we teach at. Then, we outline how this has influenced our experience of teaching upper level elective courses in Natural Language Processing and Machine Learning in previous semesters, and the changes we want to make in our proposed course as a result of this experience. 

%In this section we outline the classes that students at YYYY (the instituion we are at) take, and with it the skills and interests they are likely to bring into a 400-level Natural Language Processing class. Then, we describe the kind of NLP practitioners we want to train in this context. Next, we describe previous approaches to teaching NLP and Machine Learning at our institution and the challenges we faced. We use all of this context to motivate the learning objectives, course principle and course for our NLP class in later sections.

\subsection{Required and elective courses at Colgate}
Our curriculum has four required classes: one at the 100 level (Introduction to Computing) and three at the 200 level (Data Structures and Algorithms, Introduction to Computer Systems, Discrete Structures). In addition students have to take four electives, with at least two classes at the 300-level or above, and at least one 400-level course. All of our 400-level courses require students to work on ``capstone'' final projects.

Most electives have one or two of the required 200-level courses as prerequisites, with almost no elective requiring another elective as a prerequisite. As a consequence of this flat prerequisite structure, students can come into a 400-level elective like NLP with more limited programming experience than their peers at institutions with a more hierarchical prerequisite structure, and with likely little to no prior background in relevant classes like Machine Learning or Artificial Intelligence.  The challenge, therefore, is to design a class that sets up our students up our students, who have relatively limited programming experience and relevant conceptual background, to work on meaningful capstone projects. 

\subsection{Take-aways from previous iterations of teaching NLP and Applied ML}
In the past three semesters we've taught two sections each of NLP and Applied
ML. In all of these four classes, while all students were able to successfully
produce capstone projects, we thought most of these projects could have been
more ambitious. We've identified two factors that likely contributed to the
smaller scope of the projects: the order in which earlier non-neural approaches
vs. more contemporary neural network content was presented, and students' prior
programming experience. 

\paragraph{Order of introducing non-neural vs. neural approaches}
In both Applied ML and NLP we spent the first half of the semester working
through non-neural network approaches. For example, in Applied ML we spent the
first four weeks of the semester working through evaluation metrics, gradient
descent, regression, SVM, decision trees and random forests before introducing
neural networks. Similarly, in NLP we spent the first six to seven weeks working
through FSAs, context free grammars and parsing, n-gram language models and
Bayesian classification before introducing neural language models and transfer
learning. 

We adopted this approach because starting with the non-neural approaches makes
it easier to ground the computational task we are trying to solve with our ML
and NLP models, and the metrics we use to evaluate these models. For example, starting
with FSAs and CFGs before n-gram models highlights the complexity of modeling
structure in language and how co-occurrence statistics can capture some of this
complexity. Then, introducing classification tasks and evaluation metrics in the
context of n-gram models makes it possible to reason about the relation between
the tasks we want to solve and the metrics we use to evaluate performance on these tasks before
trying to understand the contemporary approaches that have been found to
be successful. 

A consequence of this approach, however, was that by the time we taught contemporary
approaches applicable to a wider range of tasks, it was too late for students to
effectively incorporate them into their capstone projects.  Therefore, many
students chose to use non-neural approaches in their final project which meant
that they worked on simpler tasks and/or were met with limited success on their
tasks. We propose to address this issue by using a layered approach to
introducing concepts (\S~\ref{sec:layered-approach}) and designing labs early in
the semester that scaffold some of the aspects common to all projects such as
data pre-processing and generating plots (\S~\ref{sec:lab-lecture}). 

\paragraph{Prior programming experience} 
As discussed earlier, since students can take NLP after having taken as few as
three CS courses, they can have more limited programming experience than
students at other institutions with a more hierarchical prerequisite structure that requires students to take more CS classes before taking NLP. As a consequence, we found that
students struggled with implementing more complex NLP or ML pipelines using
sparse starter code or links to existing code-bases. We propose to address this
by designing a modular toolkit with the core components required for any NLP
pipeline. The toolkit will be designed such that students can use it off the shelf
from the beginning of class. Crucially, as the class progresses and we want them
to engage with implementational details of different components, the modular
design makes it easy to ask them swap out different components that they
implement from scratch (\S~\ref{sec:toolkit}). We also plan to include a midterm
project that requires students to replicate previous work, which will give them
practice working with existing code bases and integrating it with the toolkit(\S~\ref{sec:midterm}).
% ,
% as well as giving them practice with writing ACL-style conference papers

%Additionally, we also want to add more scaffolds to the final project. We experimented with a more involved approach when teaching Applied Machine Learning that involved the following steps: individual project ideas, group project written proposal, oral feedback on proposal from two instructors, pilot presentation, devoted time to work on the project in lab, group poster presentation and individual final paper. Anecdotally, we found that on average the final projects were much better exceuted and presented than in our previous iterations of NLP. Therefore we want to adopt this approach, with some modifications (discussed in \S~REF).

% \subsection{Culture of constructive peer feedback}
% All of our classes are small, with around 18-28 students, which makes it possible for classes to have a lot of in-class small group discussions and group work where students have formal opportunities to learn from each other apart from any informal study groups they might form outside of class. This culture of constructive peer-feedback afforded by the small class sizes motivates our final project design which includes several stages of peer review [ADD A SENTENCE AND REFERENCE TO SECTION]. 

\begin{table*}
    \centering
    \begin{tabular}{llll}
     \toprule
       \textbf{Course Design Decision} & \textbf{Scholar goal} & \textbf{Tenet} & \textbf{Capstone skill}\\
        \midrule
        Layered approach & S1, S2 & T1, T3, T4 & -- \\
        Lab vs. Lecture Integration & S1 & T1, T2, T4 & C2 \\
        Toolkit & S2 & T4 & C2 \\
        Midterm replication & S2 & T3, T4, T5 & C1, C2 \\
        Capstone project & S1, S2, S3 & T4 &  C1, C2, C3 \\
        Society and Science Comm & S3 & T5,T6 & --\\
        \bottomrule
        
    \end{tabular}
    \caption{Mapping our course design decisions to the tenets and desired skills in our proposed course.}
    \label{tab:skills}
\end{table*}

\section{How to train an NLP Scholar}\label{sec:scholar}
Given the liberal arts context that we are in, and the background that students
come in with, we think that we are not very well placed to train NLP engineers
who can leave the class knowing how to design and build new tools to tackle a
wide range of NLP tasks. Rather, we are better placed to train NLP scholars who,
when given a task or challenge, can identify and apply existing tools to the
challenge or task, while thinking critically about the limitations of these
tools, our methods to evaluate them, and the societal context in which they are
used. Concretely, there are three skills we want the NLP scholars we train to
have.

%our goal is to not train NLP engineers. Instead, we want to train NLP scholars who are able to think critically about language processing, the tools used to model this, and the societal context in which these tools are used. Concretely, there are three skills we want the NLP scholars we train to have.

\paragraph{S1: Describe language processing}
NLP scholars should be able articulate clearly what computations underlie
language processing and how the systems we are building are approximating
different aspects of this at a computational and/or algorithmic level. That is,
they will develop an understanding for language as a computational system, while
setting aside concerns from psycho- or neuro-linguistics of how exactly
linguistic processing is realized. 

\paragraph{S2: Effectively using existing NLP tools}
NLP scholars should recognize what existing NLP tools are appropriate for
solving different tasks or answering different questions, and be able use these
tools to solve the tasks and answer the questions. 

\paragraph{S3: Evaluate claims about NLP systems}
NLP scholars should be able to identify the rhetorical tools used to make
arguments about the potential and limitations of NLP systems, both in academic
papers and public media, and use evidence based approaches to evaluate these
arguments.  

\subsection{Tenets of our proposed course}
Given these high level skills, we now describe some of the tenets we are adopting in this course to facilitate the acquistion of these skills. 
\paragraph{T1: Appreciate the complexity of language} Students should recognize 
%all of 
the complexity involved in language processing. We hope to accomplish this by having students create and evaluate symbolic computational models of language processing.

\paragraph{T2: Focus on multilingualism} Students should recognize the role that linguistic diversity and multilingualism plays in our understanding of language as a phenomenon, and describe the scientific and societal benefits of modeling languages other than English.  

\paragraph{T3: Recognize how NLP tasks abstract away from the complexity} Students should describe how different NLP tasks (e.g., language modeling) abstract away from the complexity of language processing. They should also be able explain the practical importance of this abstraction while articulating the limits that this abstraction poses on the conclusions we can draw about language processing from building and studying these models.

\paragraph{T4: Build NLP systems using existing tools} Students should be able to describe all of the components of building NLP systems. They should also be able to use existing code-bases and tools to design systems that solve specific tasks or answer specific questions. 

\paragraph{T5: Critically examine the role of benchmarks} 
Students should be able to articulate how and why benchmarks shape NLP research and product development, while reasoning about the limitations of benchmarks.

\paragraph{T6: Critically examine the impact of hype culture on science and society}
Critically reason about what kinds of results about NLP systems' capabilities (or their lack thereof) get hyped and why, while describing the impact that this hype can have on society. 

\subsection{Capstone specific skills}
We think that an effective way to achieve the skills is to learn by doing, and we think that engaging in a capstone project that requires students to answer a concrete question or solve concrete task is an ideal way to learn NLP. We describe the different components of our proposed capstone project in \S~\ref{sec:capstone}, but overall, we want students to be able to do the following. 

\paragraph{C1: Reading Scientific Papers} Students should be able to distill the hypotheses, methods, results and conclusions from a scientific paper while critically evaluating whether the conclusions follow from the results. 

\paragraph{C2: Replicate prior work} Students should be able to follow the procedures described in a paper to replicate prior work, and reason about what counts as a successful replication. 

\paragraph{C3: Engage in peer-review} Students should be able to give constructive criticism in a peer-review setup, as well as incorporate constructive feedback to improve their work.

\section{Course Components and Assessments}\label{sec:course}

Following our backwards design, we intentionally related course components and
assessments to the skills we want our students to acquire, the tenets
underlying our course principles, and the skills we want demonstrated in a
capstone project (see \S~\ref{sec:scholar}). We summarize these connections in
Table \ref{tab:skills}. 

\subsection{Layered approach to introducing concepts}
\label{sec:layered-approach} In the previous iterations, we adopted a largely
sequential approach to introducing concepts. We found that this resulted in a
bit of a fragmented course structure. In this iteration we want to adopt a
layered approach that introduces core concepts at multiple levels of abstraction
at different points in the course. To accomplish this, we plan to start the
class by introducing the NLP pipeline at a very high level, and then
un-blackboxing different parts as the semester progresses. This un-blackboxing
process applies to both computational concepts (e.g., when students are asked to
write context free grammars to engage with the complexity of language) as well
as to practical implementational level details (e.g., when students are asked to
implement tokenization). We describe below how we plan to use the lab-course integration and the toolkit as tools for this layered approach. 

\subsection{Labs-Lecture intergration}\label{sec:lab-lecture}

Following other natural sciences, many courses in our department have an
accompanying lab. These labs serve to provide supplemental practice with
concepts covered in the course and concrete hands-on time for programming. 
Within the context of our NLP course, we intentionally designed labs for two
broad purposes: preparing students for midterm projects (see \S~\ref{sec:midterm} 
for more details on this project) and to more deeply 
explore content not fully covered in lecture. 

In designing lab content, we are motivated by a tension we observed in earlier
versions of this course and related courses (e.g., Machine Learning): student
final projects tend to draw on current trends in the field, but knowledge of the
field's foundations help contextualize modern techniques. As a result, while
earlier material is critical, time spent on developing intuitions for
traditional concepts (e.g., context-free grammars) reduces the scope of what
students are capable of accomplishing in their final projects. For example, in
earlier versions of NLP, there were only around 2 weeks between the introduction
of transformers and the completion of earlier phases of the final project. 

%We resolve this tension by two components of the course: the lab and
%the toolkit (see Section \ref{sec:toolkit}). 

Labs resolve this tension by providing time outside of lecture for developing
skills critical to success in their final projects. This includes, a refresher
on Python (the middle of our curriculum is taught in Java and C) and labs
targeting data processing, hypothesis generation, conducting experiments,
plotting, and interpreting results. 

In their additional role as an opportunity for deeper engagement with lecture
content, labs take the form of hands-on exploration, similar to the  `scaffolded discovery' advocated in \citet{schofield-etal-2021-learning}. Concretely, consider
context-free grammars. In lecture, they will be discussed at a conceptual level
(what are they, how do they relate to language, what are ways we may use them,
etc.). In lab, students will be asked to write actual context-free grammars for
fragments of a language, following in the style of
\citep{eisner-smith-2008-competitive}. In this capacity, labs allow us to decide
when to abstract and when to go a bit deeper, without relying solely on lecture
time. 

%\begin{itemize}
%    \item As a way of prepping midterm final
%    \item other types of content
%    \item contrast with lecture, what are we abstracting and what are we going a bit deeper on 
%    \item early labs are setting them up to do the midterm, by providing the skills early, we don't need to ground every symbolic thing. early stuff is not our aim for the final projects so creates tension with what we want them to be able do in the end. the labs offer time to add skills while still allowing class to have content 
%\end{itemize}

\subsection{Toolkit}\label{sec:toolkit}

\begin{figure}
    \centering
    \includegraphics[width=\linewidth]{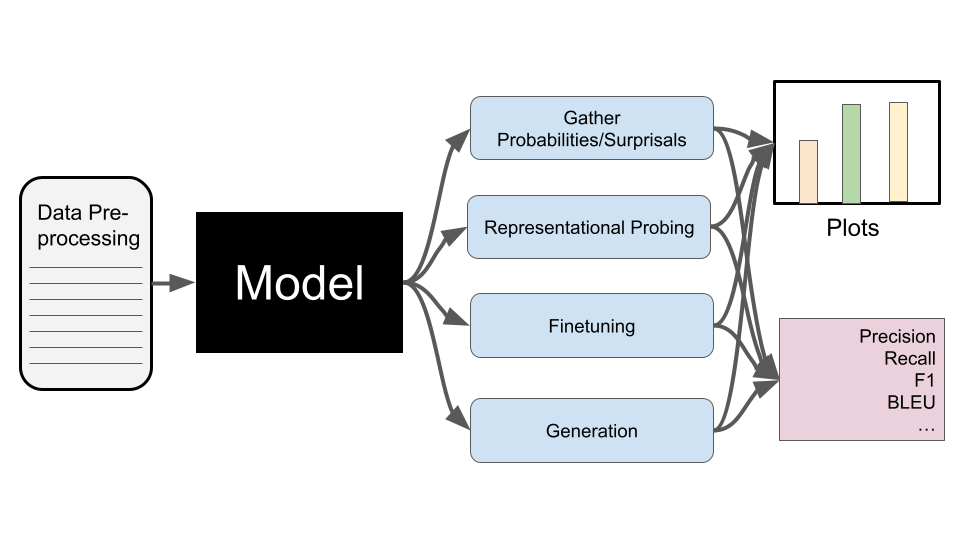}
    \caption{Sketch of the core components of the toolkit}
    \label{fig:toolkit}
\end{figure}

In developing the course, we had to decide what to blackbox and when. A variety
of factors influenced our thinking on this. We want students to develop a
high-level understanding of the core parts of an NLP pipeline and how they
relate to one another, while also leaving space for deeper dives into specific
aspects in the form of student implementations. We believe, following our course
principles, that students learn best by doing. In this case, students gain a
fuller understanding of the NLP pipeline if they can use and explore a working
implementation. Additionally, they benefit from implementing parts of this
pipeline. Our toolkit seeks to balance these two needs by being modular. 

As sketched in Figure \ref{fig:toolkit}, there are 4 basic components to our
toolkit: data pre-processing, model, experiments, and handling the output
(plotting and evaluation metrics). Data pre-processing includes both basic text
processing (e.g., reading different file types, text normalization) and
tokenization. Following an earlier implementation of this toolkit, the model
components is comprised of a parent class with basic methods (e.g., getting the
logits from a model, aligning logits to words, getting intermediate output) that
allow for a shared structure across disparate models (including transformers and
RNNs/LSTMs). The experiment portion of the toolkit facilitates basic approaches
in the field (e.g., probing, targeted syntactic evaluations). Finally, the
output of experiments interacts with both metrics (e.g., F1) and also a plotting
interface. Each of these components can be supplemented with student
implementations. For example, a lab can focus on building a different type of
tokenizer, which students can add to the pipeline to extend the capabilities of
the existing toolkit. 

In designing the toolkit we are mindful of two things: we want the code to be
readable and approachable for students to dig into while also building in a
level of abstraction that facilitates using the toolkit without fully
understanding the components early in the semester. Like in earlier versions of the
toolkit we plan to abstract from the implementation via basic plain text files
specifying experimental material (e.g., grammatical and ungrammatical sentences)
and a config file that specified pre-processing details (e.g., including
puncutation) and the desired models to run the experiment on. Additionally, we
will build on top of existing, and widely adopted NLP toolkits like NLTK and
HuggingFace (especially for models), so that students gain some familiarity with
existing industry tools. 

\subsection{Midterm}\label{sec:midterm}
\begin{table*}[t!]
\centering
%\begin{tabular}{||p{2.5cm}|p{1.5cm}|p{1.5cm}||}
\begin{tabular}{lll}
     \toprule
     Theme & Example 1 & Example 2 \\ [0.5ex]
     %\hline\hline
     \midrule
     Basic Methodology & \citet{newman-etal-2021-refining} & \citet{warstadt2020linguistic}  \\
     \hline
     Interpretability & \citet{clark-etal-2019-bert} &   \citet{hewitt-manning-2019-structural}\\
     \hline
     Experimental Design & \citet{mccoy-etal-2020-berts}&  \citet{hewitt-liang-2019-designing} \\
     \hline
     Cross-Linguistic & \citet{ravfogel-etal-2019-studying} & \citet{mueller-etal-2020-cross} \\
     \hline
     What's in the Data & \citet{yedetore-etal-2023-poor} & \citet{deas-etal-2023-evaluation} \\ [1ex]
     \bottomrule
    \end{tabular}
    \caption{Paper themes and examples. For the midterm paper, students will
    replicate papers from any category but Basic Methodology, which are used in
    lab.}
\label{tab:papers}
\end{table*}

To scaffold the final project, the middle of the semester culminates in a
midterm project, rather than a midterm exam. The intention of this project is to
provide students an opportunity to concretely apply skills relevant to the final
paper, but in a more structured format. Concretely, we identify 4 skills that we
want to ensure students can apply going into the final project: 

\begin{enumerate}

        \item Working with existing code and/or libraries \vspace{-0.5em}

        \item Hypothesis generation and operationalizing a question \vspace{-0.5em}

        \item Interpreting and synthesizing results \vspace{-0.5em}

        \item Science communication 

\end{enumerate}

Development of these skills are facilitated in lab and in the production of
their midterm paper. We believe it is challenging, if at the end of the
semester, students are faced with two tasks, i) apply new knowledge in a new
format (e.g., many other computer science courses in our department lack a final paper), and ii)
develop a novel idea that excites them. The midterm project seeks to divorce
these (to some extent). Rather than producing novel work, students are tasked
with replicating existing work. This pre-existing structure helps guide them in
the development and application of course concepts. In closely studying an
existing paper, they gain familiarity with how researchers frame questions, how
to synthesize results in a way that is digestible to the reader, and how to work
with existing code and data. 

In the following sections, we discuss how we approached selecting papers to
serve as the base of student replications, and how we scaffold the skills
necessary to complete it. 

% \subsubsection{Types of papers and why}
\paragraph{Types of papers and why}

In choosing papers, we focused on the core themes of the course and looked for
papers that were (often) short, contained existing code, exposed a key
methodology, and would facilitate good discussions. We settled on 5 themes:
Basic Methodology, Interpretability, Experimental Design, Cross-Linguistic or
Multilinguality, What's in the Data. Example papers are provided in Table
\ref{tab:papers}. The Basic Methodology theme serves a particular function in
scaffolding the midterm project, which we return to below. For the others, we
wanted to highlight different approaches for evaluating models
(Interpretability), how carefully created experiments can expose flaws in
scientific reasoning (Experimental Design), how we should think broadly about
language (cf. the ``Bender Rule'', \citealt{bender2019benderrule}; Multilinguality), and how we should think
critically about what is in our training data (What's in the Data). These
categories are not exclusive -- papers can fall in more than one category. For
example, Deas et al. (2023) invites discussion of both the contents of our
training and evaluation data, but also, highlights the diversity of what is
meant by English.  

% \subsubsection{Scaffolds for writing the paper}
\paragraph{Scaffolds for writing the paper}

In writing their midterm papers, students will demonstrate their knowledge of
creating research: going from data to question to experiments to results, and
finally a conclusion. To scaffold the acquisition of these skills, we are
relying on early labs. Students will have labs that guide them throw basic
research pipelines like formatting data to work with a model. Concretely, we
draw on papers from the Basic Methodology as an in-lab replication assignment.
Additionally, students will present their midterm projects to us for feedback,
facilitating practice with scientific communication. 

%\begin{itemize}

%        \item lab on data preprocessing, experiments, arguing for results
%        \item paper feedback 

%\end{itemize}

\subsection{Final}\label{sec:capstone}

The course culminates in a capstone project in the form of a final project,
done in small groups with individual final papers. We aim through the semester
to have assessments that help ensure the students remain on track. Concretely,
over the course of a few semesters of piloting this, we have settled on 5
phases: 

\begin{enumerate}

        \item Individual Ideation and Group Formation \vspace{-0.5em}
        \item Feedback Discussion \vspace{-0.5em}
        \item Pilot Presentation \vspace{-0.5em}
        \item Poster Presentation \vspace{-0.5em}
        \item Final Paper

\end{enumerate}

We discuss the motivation and content of each phase below. 

% \subsubsection{Ideation to Group Project}
\paragraph{Ideation to Group Project}

In driving towards a group project, we want an opportunity for students to
externalize their own interests and map it to the course content. We have an
explicit assessment to target this. Students submit an individual project
proposal that outlines their idea, a concrete link to a relevant dataset or a
description of how they would make one, a question operationalized with an
experiment or task, and their availability to work in the semester (so that
groups can take into account working style/time). See Appendix \ref{app:proposal} for more
details. Students are asked to read all the individual project ideas and to
submit a ranking of projects they would like to work on. Rather than just having
students form groups immediately, we believe this facilitates better groups by
allowing students to find overlapping interest with people they might not already
know. 

% \subsubsection{Feedback Discussion}
\paragraph{Feedback Discussion}
After forming groups, students are asked to put together a project proposal
(similar to their individual project proposals). We review these and then have
meetings with each group to help flesh out any limitations in their current
proposal. This often takes the form of helping them make their project either
more ambitious, or conversely, scaled down to something manageable in the
remaining time of the semester. As opposed to written feedback, we find these
conversations productive and useful in helping students articulate a project
that is appropriate and exciting for them. In the end of the discussion, we
establish what should comprise their pilot presentation. 

% \subsubsection{Pilot}
\paragraph{Pilot}
As the class shifts to focusing on final projects, we ensure that there is time
in labs for groups to make progress. We formalize this desire for making
progress prior to the end of the semester with a pilot presentation. Groups are
meant to implement the core part of their final project and present it to the
class in the form of a short presentation. By demonstrating that their project
works, at least in a limited capacity, groups can expose any issues that might
arise in conducting their work (e.g., the data are not good, the model they are
drawing on performs more poorly than expected). Additionally, students get
practice communicating their results to a more general audience (each other).
For each presentation, students submit feedback forms (see Appendix \ref{app:poster}) and
drive the questions in the discussion period after the presentation. 

% \subsubsection{Poster Presentation}
\paragraph{Poster Presentation}

In the final week of the semester, students create and present posters about
their final project. This ensures that they have a nearly complete final project
prior to working on the final paper during the examination period. In the past,
we have had students give feedback for each poster using a form. We have decided
to have students write anonymized reviews of a subset of posters. This, we hope,
encourages them to engage more substantially with the poster and helps teach
them the process of science (namely, peer-review). We plan to use a reviewing
form similar to that of used by ARR (e.g., highlighting the strengths,
weaknesses, and key takeaway). A copy of our poster template, rubric, and prior
feedback form is given in Appendix \ref{app:poster}. 

%\begin{itemize}
%    \item Feedback

%    \item have them write anonymized reviews of (subset of) posters 
%    * strengths
%    * weakness
%    * argument 
%    * encourages them to actually engage with poster 
%    * helps teach science approach 

%\end{itemize}

% \subsubsection{Final Paper}
\paragraph{Final paper}

Finally, each student is asked to individually write a final paper, following a
template included in Appendix \ref{app:paper}. The paper mimics a typical research paper in
NLP (drawing on ACL's style guide). Given the group nature of the other parts of
the final project, we want to retain some way to assess the individual
contributions of group members. We have found that final papers in the past have
exposed both exciting differences in interpreting results but also highlight the
varied contributions of group members. 

\subsection{Society Reflection}

A core aim of this course is not only to introduce students to NLP but also
produce critically engaged practitioners. The final component of the course is
meant to scaffold this. In their paper on incorporating reflection on social
issues in CS classes, \citet{davis2011incorporating} highlighted, among other
things, the following two challenges: First, students are good at repeating
platitudes (e.g., ``it is important to address biases in models''), but find it
challenging to critically think about the issues in practice. Second, formally
assessing students' ability to engage with these issues can be challenging. 

In our course, we hope to address these challenges by requiring students to
write a 1-2 page ``Society reflection'' paper. For this paper, students will be
asked to find and read a recent news article that discusses advances in NLP and
asked to evaluate the arguments the article makes (implicitly or explicitly)
about the societal implications of these advances. Since students are asked to
evaluate specific arguments, merely repeating platitudes will not be sufficient;
they will need to draw on their knowledge of the tools and techniques in NLP to
examine the central claims, while also reflecting on the rhetorical reasons for
making those claims. Students' ability to engage with the societal issues as NLP
practitioners will be assessed based on their ability to identify the core
claims and the soundness of their arguments when evaluating these claims.

 % In the past, students have explored data sovereignty %CITE and bias in AI. 

\section{Discussion}

% \textbf{
In this paper we presented two types of students NLP courses might want to train --- NLP engineers and NLP scholars --- and used a backwards design approach to design a capstone focused upper-level course to train the latter type of student. We started by outlining three skills we would like the NLP scholars we train to have. Then, we described six tenets and three capstone specific skills that shaped the content, organization and assessments in the course.  Finally we described these different components of the course.    
% proposed an upper level course, with a capstone experience, designed to train the latter type of student.
% } 
Having taught two sections of NLP and two sections of Machine Learning in the last two years, our proposed course represents the synthesis of many conversations, both among ourselves, but also with other educators and with our students. 

In designing the upperlevel course, we had to resolve a tension between the
knowledge and critical thinking skills we wanted our students, as NLP scholars,
to have and our desire to develop a course that fostered substantial and
exciting undergraduate NLP projects. To foster a deeper understanding and
appreciation of what NLP is about, we thought it was important to spend time on
older material (especially since we could not assume any prior background in AI
or NLP). However, to prepare our students for successful and exciting capstone
projects %, and for the workplace after graduation,%
we thought it was important to introduce them to newer tools and techniques early on. 

We resolved this tension in our proposed course in the following ways.
\begin{enumerate}
    \item We adopted a layered approach to introducing concepts where we introduce the NLP pipeline at an abstract level in the beginning, and peel away the layers as the course progresses. To accomplish this, we proposed a modular toolkit and and intentional course-lab integration. 
    \item We proposed a midterm replication project that provides students with an opportunity to work on larger scale projects with existing code-bases, while also giving them practice with critical thinking (e.g., reasoning about what counts as a successful replication) and science communication. 
    \item We proposed a highly scaffolded approach to their final capstone project involving five stages. We piloted this five-stage process in two sections of Applied Machine Learning this semester, and found that on average the student projects were more ambitious and successful. We include course materials from this pilot in the appendix. 
\end{enumerate}

% In designing the course, we had to
% resolve a tension between covering new content (with some gaps in our prerequiste structure limiting how much technical material we could assume) and
% spending time on older material that we believe facilitates a deeper understanding and appreciation of what NLP is about.  

% Additionally, we wanted to develop a course that fostered substantial and exciting undergraduate NLP projects, both to address the needs of our departments emphasis on capstone projects, but also to support students in
% building skills that will help them after graduation. To approach these goals in a principled manner, we focused on the type of student we wanted to train, emphasizing areas our students excel in (e.g., building connections across fields, engaging critically with course concepts) -- the NLP Scholar. 

% \subsection{NLP Scholar in Different Institutions and Programs}

\paragraph{NLP Scholar in Different Institutions and Programs}

While the course components we proposed are specifically designed with the constraints of our institution and program in mind, we believe aspects are applicable to training NLP Scholars in other contexts. Specifically, there are two educational environments that we would like to highlight: a computer science department at a research heavy (i.e., R1) university and other departments like linguistics or psychology where a computational linguistics course would be relevant. For both, the toolkit and societal reflection seem appropriate and useful. %However, we believe the differing demands of these contexts necessitate careful treatment of the remaining component of the course -- the final project structure and the lab/lecture. 

The final project, as we have constructed it, requires ample one-on-one time with students, both for formal components of the grade (e.g., the project feedback and in-class presentations), but also, for helping student groups progress in their project (e.g., debugging issues). 
%Additionally, we include time for in-class presentations. 
We believe these would be difficult to scale in R1 settings without relying
heavily on TAs and restructuring the progression of the final projects. However,
we believe the final project structure, and in particular, the replication
component, would be applicable across disciplines. Having linguistics students
replicate work on targeted syntactic evaluations, for example, is a good way of
taking a common practice in linguistics (the construction of minimal pairs) and
applying it to a computational domain. On the other hand, we think the
lab/lecture structure is well suited to R1 computer science students, while it
may pose challenges in being ported to other disciplinary backgrounds. We are
assuming that students in our class need more time to the concepts than on the aspects of programming. 

\paragraph{Usefulness of the NLP Engineer vs. NLP Scholar distinction} 
Even if the goal of many NLP courses might be to train students who are a
combination of an NLP Engineer and Scholar, we think the distinction can be
useful when faced with conflicting goals for the course. For example,
\citet{schofield-etal-2021-learning} advocated for a discovery based approach to
teaching NLP, but concluded that students might need more guidance on/practice with aspects like File I/O, loading and saving data, manipulating numpy
and spaCy objects etc. Here, the NLP Engineer vs. Scholar distinction can help
instructors decide what components should students productively struggle with,
and which components they are better of getting more structured guidance on.

If
the goal of the course or the specific assignment is to equip students with the
ability to \textit{build} something like an NLP Engineer, then it might be
helpful for students to productively struggle with components like File/IO with
minimal guidance. If on the other hand, the goal is to have students
\textit{use} a tool to answer a question or explore a topic like an NLP Scholar,
then students might be better off having more guidance on components of the
assignment like File I/O, so they can devote more time to the exploration and
analysis. This line of reasoning is applicable even for graduate level NLP
courses or undergraduate courses where students come in with much stronger programming experience, and are unlikely to be challenged by programming aspects like File I/O --- in these courses, instructors will need to decide the extent to which other implementational details such as GPU parallelization should be scaffolded.

\paragraph{Conclusion}
We differentiated between two types of students --- NLP scholar and NLP engineer --- and proposed an upper level course, with a capstone experience, designed to train the latter type of student.
Ultimately, we hope that articulating the goals for an NLP scholar, and tying them back to specific components in our proposed course can facilitate broader discussions about what kinds of NLP practitioners we want to train, why, and how best to train them. 

\section*{Acknowledgments}

We would like to thank our colleagues in computer science and our students over the years for helping shape the course. In particular the students in NLP at Colgate in Spring 2023 and Fall 2023, the students in Applied Machine Learning at Colgate in Spring 2024, and the students in the Methods in Computational Linguistics seminar at MIT in Fall 2022. Additionally, we would like to thank our anonymous reviewers for their helpful comments and suggestions.  

% Entries for the entire Anthology, followed by custom entries
\bibliography{anthology,custom}
\bibliographystyle{acl_natbib}
\newpage
\appendix
\onecolumn

\section{Project Proposal and Pilot Presentation Templates}
\label{app:proposal}

Students are asked to make two project proposals. One that represents their
individual idea and interest, and then later, a proposal for a group project. A verison of the proposal template is copied below. 

\subsection{Proposal Template}

\subsubsection{Introduction/ Motivation}

\begin{itemize}
    \item What is the big picture question you are trying to answer/ problem you are trying to solve? Why is this an interesting and worthwhile question to answer/ problem to work on? 

    \item What is the specific question you will pursue? Why? (Note: you will need to pick something that is feasible to answer in 3-4 weeks)

    \item What are all the possible outcomes of your project? Do you think one (or few) of these outcomes are more likely than others? Why? 

\end{itemize}

\subsubsection{Background/Literature review}

Find at least three papers related to your project. For each project write a paragraph or two summarizing: 

\begin{itemize}
    \item What were the goals of that paper? How is it related to your project? 

    \item What methods did the paper use? 

    \item What were the conclusions? 

\end{itemize}

Google scholar (or other comparable database search) is a better place to look than standard search: you are less likely to find blogposts with unverified content on Google scholar. Note, on Google Scholar you might see a lot of preprints from arxiv, even if they were also published elsewhere. It is good practice to try and find the most recent version of a paper. The general rule of thumb you should use: peer reviewed published papers are more credible than preprints. 

\subsubsection{Planned methods}

Describe what methods you plan to use to address your question and describe how your methods compare to prior work you describe in the background section. 

\begin{itemize}
    \item What dataset will you use? Is it already available or do you have to create it? 

    \item What model(s)/approaches will you use? 

    \item How will you evaluate your models? What counts as success? What conclusions can you draw (if any) if you get negative results? 

\end{itemize}

\subsubsection{Proposed timeline/division of labor}

Breakdown your project into separate tasks. For each task, list the expected amount of time, who plans to work on it and when they expect to complete it by. For this part, it might be most straightforward to fill out a table following the format below.

\textbf{One of your tasks should be “Prepare for pilot result presentation” and your timeline should highlight what work you hope to accomplish before the pilot results.}

% \begin{tabular}{|p{0.9cm}|p{1.2cm}|p{1.6cm}|p{1cm}|}
% \begin{tabular}{|p{0.9cm}|p{1.2cm}|p{1.6cm}|p{1cm}|}
%     \hline
%     \textbf{Task} & \textbf{Time required} & \textbf{Expected date of completion} & \textbf{Person} \\
%     \hline
%      & & & \\
%      \hline
% \end{tabular}

\begin{table}[h]
    \centering
    \begin{tabular}{llll}
    \toprule
      \textbf{Task} & \textbf{Time required} & \textbf{Expected date of completion} & \textbf{Person} \\
    \midrule
     & & & \\
     & & & \\
     % & & & \\
     \bottomrule
    \end{tabular}
    %\caption{Caption}
    \label{tab:my_label}
\end{table}

\vspace{1cm}

\subsection{Pilot}

After discussions between us and each group, students work towards a final
project pilot presentation. This presentation demonstrates early progress on
their project and showcases their question and operationalization. It takes the
form of a 5-8 minute presentation with time for questions from the other
students. During the presentations, students are asked to provide:

\begin{itemize}
    \item Feedback about the content (question, methods, result interpretation, conclusion etc)

    \item Feedback about the presentation (framing, visuals, oral presentation etc)

\end{itemize}

\section{Poster Presentation Templates}\label{app:poster}

At the conclusion of the semesters, groups create and present a final poster. Details on the poster template, grading rubric, and student feedback from are provided below.

\subsection{Poster Template}

\begin{figure*}[h]
    \centering
    \includegraphics[width=\linewidth]{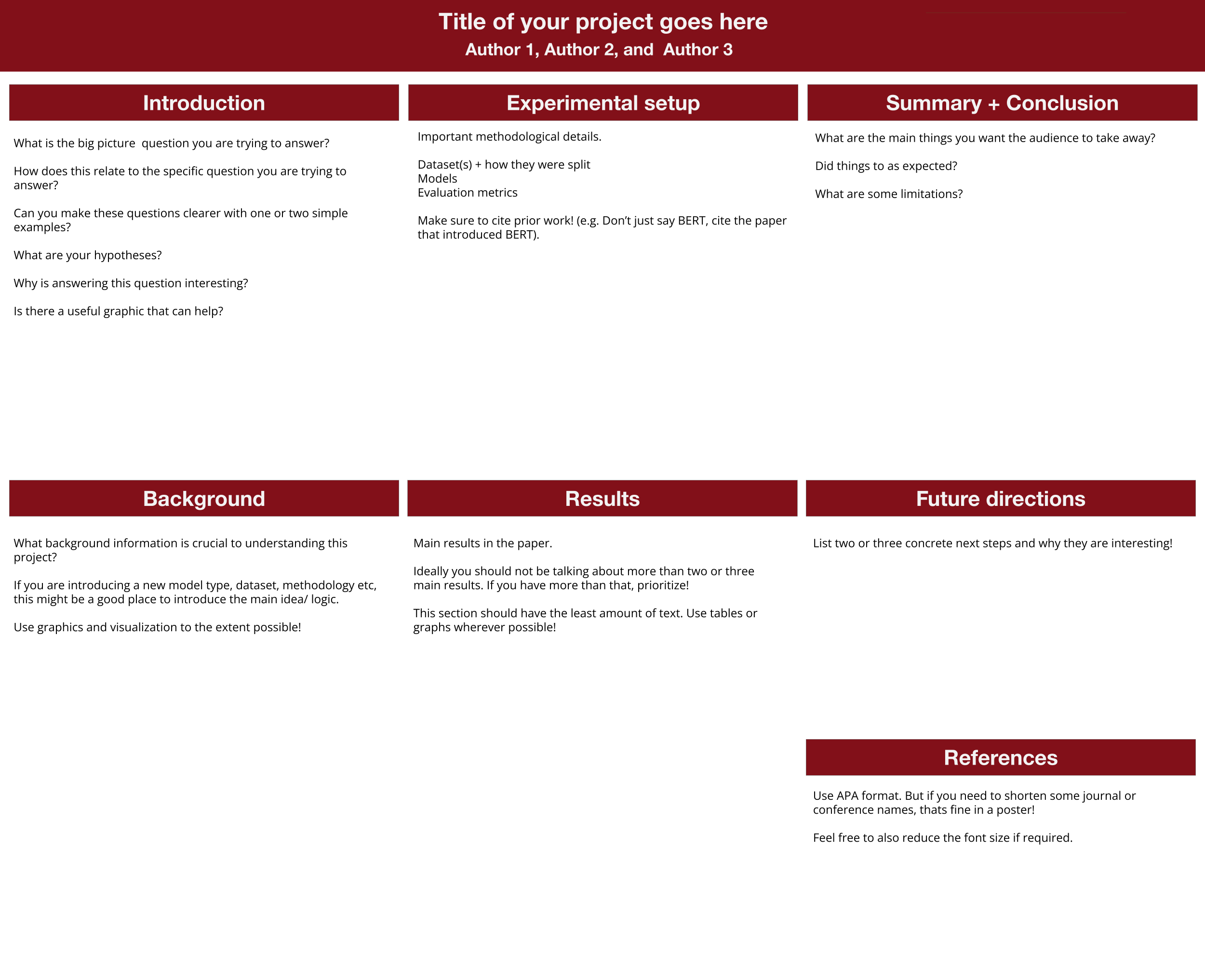}
    \caption{Final poster template}
    \label{fig:poster}
\end{figure*}

We provide students with a final poster template (given in Figure \ref{fig:poster}) in the form of a google slide and are asked to make use of the on-campus printing services to gather their physical poster. 

\subsection{Poster Rubric}

In an effort to balance listening to and grading the presentations, we settled on a brief rubric in Table \ref{tab:posterRubric}. 

\begin{table*}[h]
    \centering
    \begin{tabular}{|p{11cm}|p{1cm}|p{1cm}|p{1cm}|}
        \hline
        \multicolumn{4}{|l|}{\textbf{Extractable from poster?} (0: missing      1: yes      2: clear)} \\
         \hline
         Introduction & 0 & 1 & 2 \\
                  \hline

         Background & 0 & 1 & 2 \\
                  \hline

         Experimental Setup & 0 & 1 & 2 \\
                  \hline

         Results & 0 & 1 & 2 \\
                  \hline

         Summary + Conclusion & 0 & 1 & 2 \\
                  \hline

         Future work & 0 & 1 & 2 \\
         \hline
        \multicolumn{4}{|l|}{\textbf{Clarity of presentation} (0:  bad         1: ok          2: exceeds expectations) } \\
        \hline
        Division of labor (ok: majority of team contributes) & 0 & 1 & 2 \\
                  \hline
        Oral pitch (ok: leave with question, operationalization, takeaways) & 0 & 1 & 2 \\
                  \hline
        Timing (ok: <= 5 mins) & 0 & 1 & 2 \\
                  \hline
        Visuals (ok: informative, legible) & 0 & 1 & 2 \\
                  \hline
        Answering Questions (ok: understand the question ask) & 0 & 1 & 2 \\
                  \hline

    \end{tabular}
    \caption{Final poster rubric}
    \label{tab:posterRubric}
\end{table*}

\subsection{Student Feedback on Posters}

When not presenting, students are asked to engage with the other presentations and provide feedback. In prior courses, this feedback too the form of filling out a paper form. Concretely, they were asked to provide (at least) one bit of constructive feedback along each of these dimensions: 

\begin{itemize}
    \item Content
    \item Visualizations
    \item Argument presentation
    \item Future direction
\end{itemize}

\section{Final Paper Templates}\label{app:paper}

Using a latex template styled on ACL's submission template, students are asked to write an individual final paper. Some guidelines we gave to students are provided below. 

\subsection{Final Paper Guidelines}

Your final paper should have the following sections. 

\subsubsection{Introduction/Motivation}

\begin{itemize}
    \item What is the big picture question you are trying to answer/ problem you are trying to solve? Why is this interesting? 
    \item What is the specific question you are pursuing? Why?
\end{itemize}

\subsubsection{Background/Literature review}
Find at least three papers related to your project. For each paper write a paragraph or two summarizing: 

\begin{itemize}
    \item What were the goals of that paper? How is it related to your project? 
    \item What methods did the paper use? 
    \item What were the conclusions? 

\end{itemize}

Note, on Google Scholar you might see a lot of preprints from arxiv, even if they were also published elsewhere. It is good practice to try and find the most recent published version (i.e. conference version) of a paper. 

\subsubsection{Methods}
Describe what methods you used to address your question and describe how your methods compare to prior work you describe in the background section. 

\begin{itemize}
    \item What dataset did you use? Was it already available or did you have to create it? 

    \item What model(s)/approaches did you use? 

    \item How did you evaluate your models? What counted as success? What conclusions can you draw (if any) if you got negative results?

    \item Include a link to a repository with your code (or similar) in the paper

\end{itemize}

\subsubsection{Results}
Describe what you found in your work. Put your results in a figure or a table that helps the reader synthesize what you’ve done.

\subsubsection{Discussion}
How does your work answer your question? What are the implications of your results? What are ways the work could be extended? What are the limitations of your work?

\end{document}